\documentclass{article}

\usepackage{PRIMEarxiv}

\usepackage[utf8]{inputenc} 
\usepackage[T1]{fontenc}    
\usepackage{hyperref}       
\usepackage{url}            
\usepackage{booktabs}       
\usepackage{amsfonts}       
\usepackage{nicefrac}       
\usepackage{microtype}      
\usepackage{lipsum}
\usepackage{fancyhdr}       
\usepackage{graphicx}       
\graphicspath{{media/}}     

\usepackage{algorithm}
\usepackage{algorithmic}
\usepackage{amsmath} 
\usepackage{amssymb} 

\title{Bilinear-Convolutional Neural Network Using a Matrix Similarity-based Joint Loss Function for Skin Disease Classification
}

\author{
Belal Ahmad$^{\mathrm{\ast}}$\thanks{$^{\mathrm{\ast}}$School of Computer Science and Technology, Huazhong University of Science and Technology, China} \\
  \texttt{belalamu.63@yahoo.com} \\
   \And
Mohd Usama$^{\mathrm{\dagger}}$\thanks{$^{\mathrm{\dagger}}$Departments of Diagnostics and Intervention, and Biomedical Engineering, Umea University, Sweden} \\
  \texttt{mohd.usama@umu.se} \\
   \And
Tanvir Ahmad$^{\mathrm{\ddagger}}$\thanks{$^{\mathrm{\ddagger}}$School of Information and Control Engineering, Xi’an University of Architecture and Technology, China} \\
  \texttt{tanvir@xauat.edu.cn} \\
   \And
Adnan Saeed$^{\mathrm{\S}}$\thanks{$^{\mathrm{\S}}$School of Civil and Hydraulic Engineering, Huazhong University of Science and Technology, China} \\
  \texttt{adnansaeed@hust.edu.cn} \\
     \And
Shabnam Khatoon$^{\mathrm{\P}}$\thanks{$^{\mathrm{\P}}$School of Management Science and Engineering, China University of Geosciences, China} \\
  \texttt{shabnamali.ali9@gmail.com} \\
     \And
Long Hu$^{\mathrm{\ast}}$\thanks{$^{\mathrm{\dagger}}$Work performed while at School of Computer Science and Technology, Huazhong University of Science and Technology, China} \thanks{$^{\mathrm{\dagger}}$Corresponding author} \\
  \texttt{hulong@hust.edu.cn}\\
}

\makeatletter
\renewcommand{\@fnsymbol}[1]{\relax}
\makeatother

\begin{document}
\maketitle
\setlength{\headheight}{22.37994pt}

\begin{abstract}
In this study, we proposed a model for skin disease classification using a Bilinear Convolutional Neural Network (BCNN) with a Constrained Triplet Network (CTN). BCNN can capture rich spatial interactions between features in image data. This computes the outer product of feature vectors from two different CNNs by a bilinear pooling. The resulting features encode second-order statistics, enabling the network to capture more complex relationships between different channels and spatial locations. The CTN employs the Triplet Loss Function (TLF) by using a new loss layer that is added at the end of the architecture called the Constrained Triplet Loss (CTL) layer. This is done to obtain two significant learning objectives: inter-class categorization and intra-class concentration with their deep features as often as possible, which can be effective for skin disease classification. The proposed model is trained to extract the intra-class features from a deep network and accordingly increases the distance between these features, improving the model's performance. The model achieved a mean accuracy of 93.72\%.
\end{abstract}


\section{Introduction}
Skin cancer is one of the most widespread diseases due to direct exposure to ultraviolet radiation and viruses. According to the statistical survey, 106110 adults (62260 men and 43850 women) were diagnosed with melanoma last year in the US, whereas above 3.3 million non-melanoma cases such as basal cell carcinoma and squamous cell carcinoma are tested amongst the 5.4 million individuals in 2018\cite{rogers2015incidence}. The rates of new occurrences of skin cancer are expected to increase every year\cite{Siegel2015cancer}. Recently, the research analysis shows that 53\% of new melanoma cases have been increased to diagnosed every year in the last decade. According to this analysis, 20\% of Americans are affected by skin cancer during their life\cite{robinson2005sun}. However, the survival rate of skin cancer can be quite promising by early diagnosis using appropriate treatment, or a five-year survival rate can be reduced from 99\% to 14\% \cite{dorj2018skin}. Furthermore, the new cases of non-melanoma have increased by up to 77\% in diagnosis from the last two decades. Basal cell carcinoma is the main reason behind the death of 3000 people every year\cite{mohan2014advanced}. Thus, early detection of different types of skin cancer is a significant task to prevent it from getting worse and a chance for better diagnosis\cite{nikolaou2014emerging}. Human vision is often subjective, which is hardly reproducible with low precision\cite{morton1998clinical}. Additionally, dermoscopy is an effective technique for capturing a high-resolution image that enables dermatologists to identify features that the naked eye cannot see. Through many traditional methods, attain better results in the diagnosis of skin cancer over the eye inspection method\cite{salerni2013meta}. This approach comes to inexperienced medical domains, which would result in poor performance and be time-consuming, even based on expert judgment. Therefore, due to being extremely subjective, it can give different diagnostic results in many cases. Skin disease classification is very challenging due to the visual similarity between malignant skin tumors and non-cancerous skin lesions, which continually transform with the occurrence of certain areas of the skin. The average reported sensitivity is not adequate for skin cancer detection, even amongst expert dermatologists. Therefore, automatic skin cancer classification is crucial in public health to obtain a valuable result. 

To solve this problem, many diagnostic systems based on a handcrafted feature learning approach obtained valuable results in melanoma detection\cite{ahmad2020discriminative, fabbrocini2014automatic, moturi2024developing}. Still, these methods are ineffective in implementing many skin disease image categories. Hand-crafted feature learning approaches are useful for single or small classes of skin disease images. Furthermore, the hand-crafted feature learning algorithms are an impractical cause of the variation of the natural aspect of skin diseases\cite{saez2014pattern}. To tackle the problem, select the essential features by feature learning instead of feature engineering\cite{bengio2013representation, usama2020self}. Although, many feature learning-based classification methods have been suggested\cite{chang2015stacked, arevalo2015unsupervised, usama2018deep}, emphasizing the mitosis diagnosis that indicates cancer and is confined to dermoscopy or histopathology\cite{wang2014cascaded, de2024dermoscopy}. To overcome this problem, we proposed a discriminative feature learning-based method for skin disease classification. The contributions of this paper are as follows:
\begin{itemize}
\item Proposed a BCNN-based model for skin disease classification using a CTN.
\item Used CTN and Xception model to learn discriminative features from skin disease images. The CTN uses a matrix similarity-based joint loss function that regularizing the weight vectors is essential to increase the distance between highly correlated subcategories of images. In contrast, the classical loss function optimizes the sample features and closes them to the weight vector.
\item To the best of our knowledge, We are the first to use matrix-similarity-based joint loss function for skin disease classification.
\end{itemize}

\section{Related Work}
The robust feature extraction of objects is a crucial part of image classification. The traditional classification methods mainly emphasize part detection to develop correspondence between object instances and minimize the impact of visual variations for an object in a strongly supervised condition. In order to apply in practical uses, many researchers have started to analyze how to correctly find the discriminative spaces and extract features using a CNN for these regions under weakly supervised conditions. Fu et al.\cite{fu2017look} proposed a recurrent attention-based recursive learning method for representing the discriminative features-space on a different scale; however, this approach considered the computational cost. In this method, CNN ignores the global information by directly detecting components of discriminative regional objects. 

Xiao et al.\cite{xiao2015application} presented a weakly-supervised classification approach based on two-level attention, whenever object-level attention chooses an appropriate bounding box of a specific object, although component-level attention places the discriminative features of the object. This approach has obtained an accuracy of about 70\% on the CUB-2000-2011 dataset. Lin et al.\cite{lin2015bilinear} proposed a B-CNN-based method using two feature extractors: first, acquires local features of the full image, and second extracts global features by pooling over regions to make an alternative descriptor for classification. However, the computational complexity of this method is too high. Kong et al.\cite{kong2017low} designed a low-rank approximation method by using a traditional B-CNN with adding a weakly supervised localization, which tried to avoid direct computation for the final product of the covariance matrix for reducing the impact of background interferences to accurate feature extraction. A common phenomenon for feature learning has been suggested during image classification in case of variation between different class objects. Additionally, their feature dimension is further reduced. Although this low-rank approximation and the original work of Lin et al.\cite{lin2015bilinear} directly apply the input, whereas many background interferences are available in input data, especially for a small target. Chopra et al.\cite{chopra2005learning} presented the Siamese network that differentiates pairs of images from different classes. The method requires the distance between different class images to be larger than the distance between the same class images with a certain margin. Esteva et al.\cite{esteva2017dermatologist} designed a universal method for skin disease classification. The method fine-tuned two pre-trained networks, VGG16 and VGG19 for feature extraction. The method achieved 60.0\% and 80.3\% Top-1 and Top-3 classification rates respectively, that significantly surpasses the interpersonal ability in their work and encourages using the related approach to get a more effective result.

Tajbakhsh et al.\cite{tajbakhsh2016convolutional} designed a transfer learning-based method. The method shows that training a pre-trained model is more effective than CNN on scratch for a limited amount of input data. The method fixed the problem of training a pre-trained network on skin disease images from different medical domains and obtained better results than training the deep CNN from the beginning. Han et al.\cite{han2018classification} designed a method for skin cancer classification, which fine-tunes the Microsoft ResNet-152 model using three datasets that consist of 12 categories of skin disease images. The method achieved 87.1\% – 6.0\% sensitivity. German et al.\cite{capdehourat2011toward} proposed a method to diagnose skin cancer, which also discussed how to use Ada Boost MC separately using skin lesion images from different categories to detect cancer. Daniel et al.\cite{ruiz2011decision} designed several methods for skin disease classification like multilayered perceptron, K nearest neighbors algorithm, and the Bayesian classifier, which analyze the images of skin disease intended for determining the degree of damage using characteristics of the affected area, which is useful to decide for extracting those parts. Additionally, evaluated independently and together to make a collaborative decision support system. The method achieved around 87\% classification accuracy. Amelard et.al.\cite{amelard2014high} proposed a method to identify melanoma by proposing intuitive features (HLIFs). These features are designed to model the ABCD standard generally by dermatologists, which represents the human-observable feature. Since the intuitive diagnostic purpose can be conveyed to the user, the method improves the classification accuracy by combining the proposed HLIFs with a set of low-level features and shows that HLIFs separate the data more effectively than low-level. Saez et al.\cite{trigueros2018enhancing} proposed a method using an essential part of input images to improve the result over traditional methods, which suffer the degradation in performance due to partial occlusions. Additionally, it introduced a loss function (batch triplet loss) using a new term to improve the performance of triplet loss. This method leads to minimizing the standard deviation of positive and negative records by using deep CNN.

\section{Background Knowledge}

\subsection{Xception}
\begin{figure*}
\centering
{\includegraphics[scale=0.60]{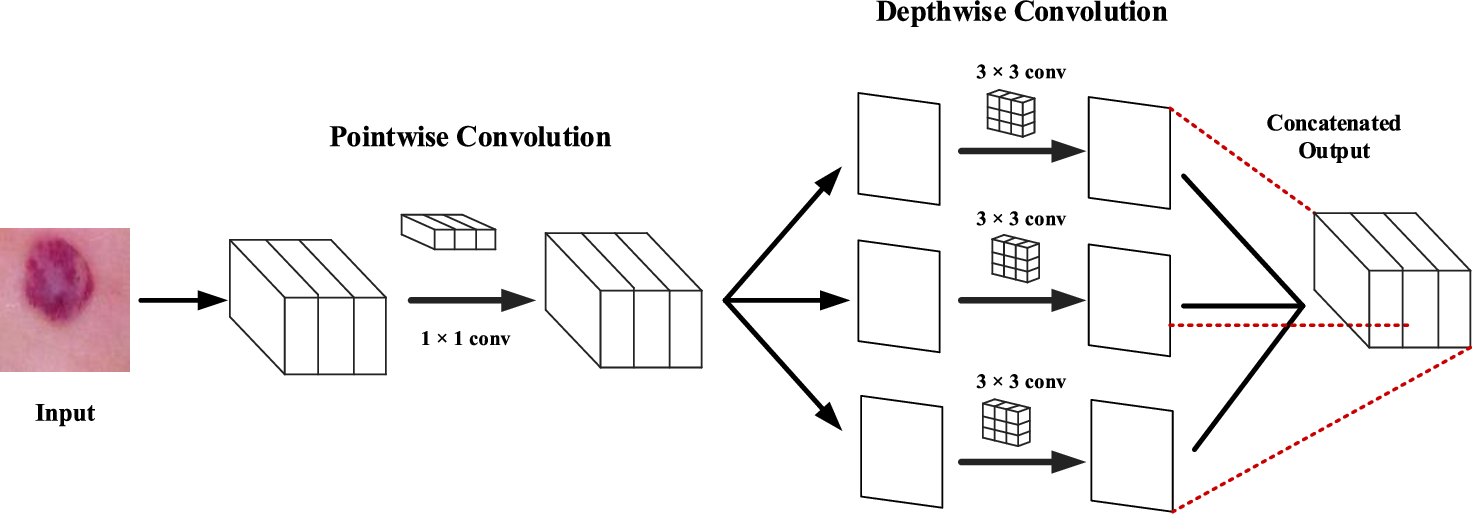}}
\caption{ The extreme version of strictly equivalent reformulation of the simplified Inception module, which has all the output channels of $[1 \times1]$ convolution with a spatial convolution.}
\label{Figure1}
\end{figure*}

The xception model\cite{chollet2017xception} is an extreme version of InceptionV3\cite{szegedy2016rethinking} that uses depth-wise separable convolution as shown in Figure \ref{Figure1}. Traditional CNNs have a deep convolution kernel and the convolution layer, which creates correlations between depth and space. The Xception model completely separates spatial correlation with a cross-channel correlation that is not mapped together to the best case. Every single channel output is individually mapped with spatial correlation; then, the cross-channel correlation is obtained using a 1×1 convolution instead of the input data being divided into several compressed data chunks. A 3D map consists of a 2D and 1D map that separately performs spatial convolutional for each channel followed by a 1×1 convolution channel, which will be assumed to request the first correlation through a 2D space and then request the correlation through 1D space. The model width increases because of separable convolution, improving recognition accuracy and enhancing the model's ability to learn sensitive and complex features.

The input image was resized to (150, 150, 3), normalized using Xception on ImageNet, and then fine-tuned on skin disease image datasets. We explored several pre-trained models and selected the Xception model as a basic model after comparing the results, since the Xception model achieved better results than every implemented model, which indicates the Xception model extracts more complex and sensitive features in the proposed method.

\subsection{Bilinear Convolutional Neural Network}
\begin{figure*}
\centering
{\includegraphics[scale=0.65]{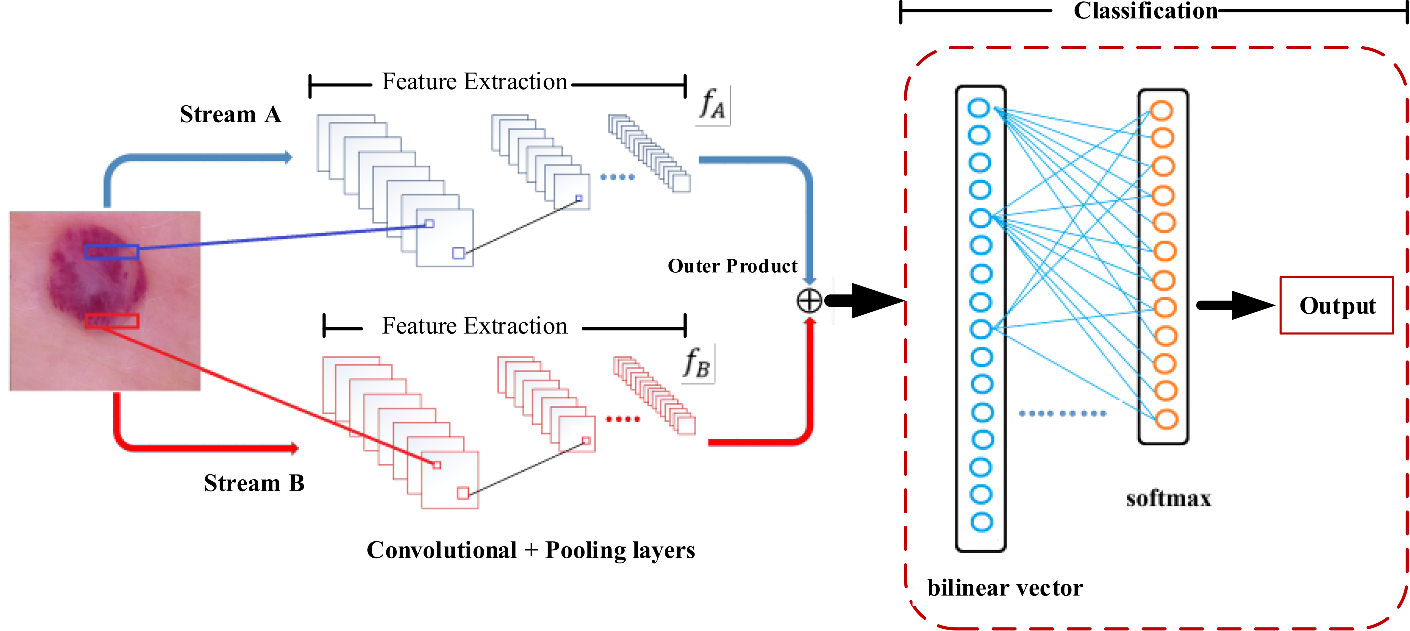}}
\caption{Streams A and B of BCNN extract the features of inputs and use an outer-matrix product to combine the outputs at each location. Then, obtain the bilinear feature representation using average pooling. At the last bilinear feature, representation passes through the softmax layer to class prediction.}
\label{Figure2}
\end{figure*}

In this method, we used a BCNN for skin disease classification that consists of quadruples $B = (f_A, f_B, P, S)$ as shown in Figure \ref{Figure2}; where  $f_A$ and $f_B$ are functions used for feature extraction, $P$ is the pooling function, and $S$ is a classification function that computes the similarity between different class images. The function for feature extraction takes an image $i\epsilon l$ and a location $l\epsilon L$  and maps a vector notation of features as $f:l\times L\rightarrow R_{(Y\times D)}$ of size $Y\times D$. We usually recommend only locations that consist of the position and scale. The outer-matrix product combines feature vector notation at each location, i.e., for any location l, the bilinear combination of $f_A$, and $f_B$ is given by Equation \ref{eq1}:

\begin{equation} bilinear(i, l, f_A, f_B ) = f_A (i, l)^T f_B (i, l) \label{eq1}\end{equation}

It is essential to appropriate feature representation for a specific model, which requires both $f_A$ and $f_B$ needs to be equal feature dimension $Y$. The pooling function generates the global feature representation $\Phi(I)$ of an image I by accumulating the bilinear feature combinations at every location of an image. In our method, we used pooling sum as shown in Equation \ref{eq2}:

\begin{equation} \phi (I) = \sum_{l \epsilon L} bilinear(i, l, f_A, f_B) = \sum_{i \epsilon l} f_A(i, l)^T f_A (i, l) \sum_{l \epsilon L} f_B(i, l)^T f_B (i, l) =  \sum_{i \epsilon l} \sum_{l \epsilon L} f_A(I, l), f_B(I, l)^2 \label{eq2}\end{equation}

where $\sum_{i \epsilon l} \sum_{l \epsilon L} (f_A(I, l), f_B(I, l))^2$ 
symbolizes a vector projection with low dimensions, the global feature representation $\Phi(l)$ for an image is order-less due to ignoring the feature locations by pooling, hence, for every spatial location $l$ in a location space $L$ of image $i$, we compute the bilinear feature vector. If $f_A$ and $f_B$ extract corresponding local feature vectors of size $Y\times N$ and $Y\times M$, then their bilinear combination $\Phi(l)$ will have size $N\times M$. The similarity function S uses the general-purpose bilinear vector representation. Instinctively, the outer product indicates the pairwise interactions of the feature extractor $f_A$ and $f_B$ on each other.

\subsection{Constrained Triplet Network}
Nowadays, the DL-based methods using TLF become more prominent due to interacting with the performance skill with extreme labels. The multi-label classification using CNN linearly increases the number of parameters. Song et al.\cite{oh2016deep} designed an N-way softmax classifier with an extreme level of labels, which used a CNN with triplet loss function by learning compact embedding to get the effective classification result. Although TLF learns good embedding $f(I^i )$ from an input image Ii into d-dimensional feature space $R^d$, i.e., $f(I^i )\epsilon R^d$ then compute $L2$ distance between each input image. Moreover, it induces the particular image $I_a$ (anchor) closer to the same category images $I_p^i$ than different category images $I_n^i$ as an objective of this method is shown in Equation \ref{eq3}:

\begin{equation} ||f(I_a) - f(I_p^i)||^2_2 + \alpha_t < || f(I_a) - f(I_n^i)||^2_2 \label{eq3}\end{equation}

where $f(I_a )$, $f(I_p^i )$ and $f(I_n^i)$ are the embedding of a triplet $(I_a, I_a^i, I_n^i)$ from the set of all triplets $T$ with major $N$ and threshold $\alpha_t$ is a pre-defined margin that imposes between images from different categories. The $L2$ distance minimizes the triplet loss value, which is defined as:

\begin{equation} L_t = ||f(I_a) - f(I_p^i)||^2_2 - ||f(I_a) - f(I^i_n)||^2_2 + \alpha_t  \label{eq4}\end{equation}

All triplets cannot be active during training which avoids slower convergence. Thus, it is essential to choose hard triplets to activate during training to improve the result. For this, every input image from the same class is projected at a single point in Euclidean space. However, triplet loss segregates the pair images from the same class (positive) and different class (negative) from a particular class image with an extra margin. That means the additional margin ensures that the images from the same category (same disease) stay within the same cluster now. It will be used to differentiate identities. 

\begin{figure*}
\centering
{\includegraphics[scale=0.40]{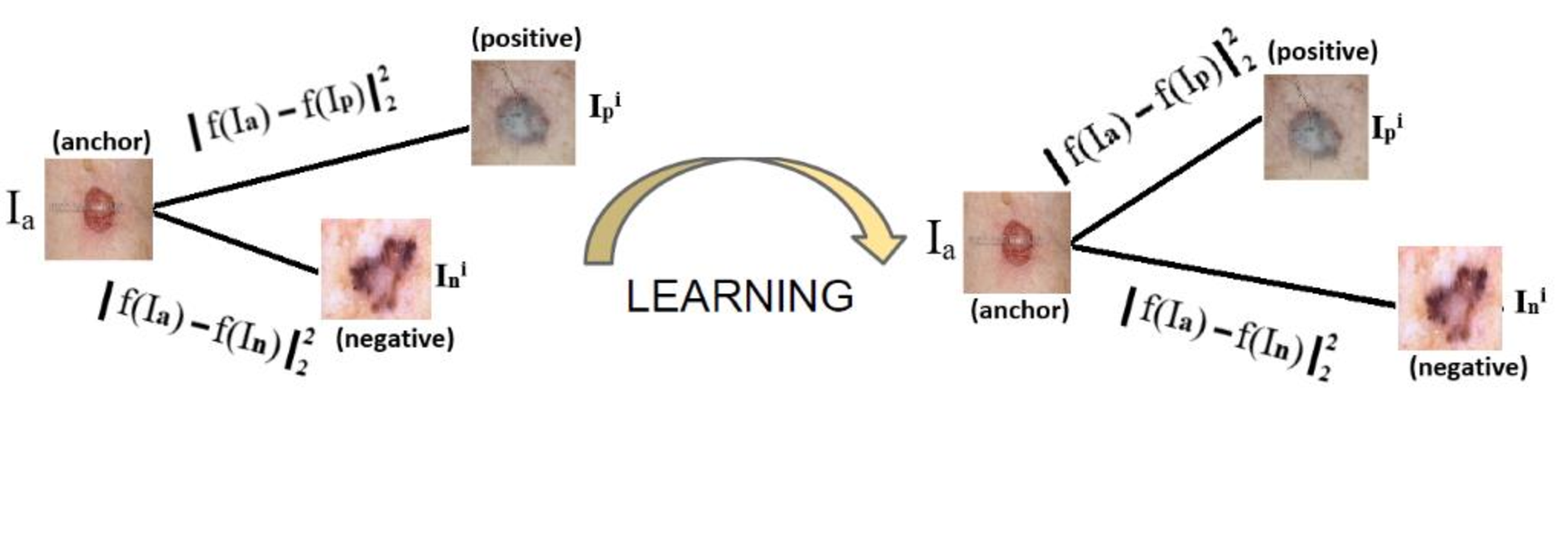}}
\caption{ Triplet loss increases the distance between skin disease images of different categories and decreases the distance between the same categories of skin disease images.}
\label{Figure3}
\end{figure*}

Furthermore, the loss is discussed from the perspective of the KNN\cite{schroff2015facenet}. The embedding is represented as $f(I^i)$, which means the embedding of each disease image $I_i$ in d-dimensional Euclidean space $R^d$, i.e., $f(I^i )\epsilon R^d$. A triplet consists of three images; $I_a$, $I_p^i$, and $I_n^i$; $I_a$ indicates an anchor input image, whereas the specific images from a positive and negative class are termed as $I_p^i$ and $I_n^i$, respectively. As mentioned earlier, the TLF distributes the d-dimensional Euclidean space in the form of a cluster of the individual class of images. To obtain a better map of this method, TLF increases the distance from the one-class image to a different class, as shown in Figure \ref{Figure3}. Mathematically, we are going to enforce a predefined margin $\alpha_t$ between $||f(I_a )-f(I_p^i )||_2^2$ and $||f(I_a )-f(I_n^i )||_2^2$ is defined in Equation \ref{eq5}:

\begin{equation} ||f(I_a) - f(I_p^i)||^{2}_{2} - || f(I_a) - f(I_n^i) ||^{2}_{2} > \alpha_t \label{eq5}\end{equation}

where $f(I_a)$, $(I_p^i)$ and $f(I_n^i)$ are the embedding of $I_a$, $I_a^i$, and $I_n^i$ from a set of triplets $T$ with cardinality $N$, $f(I_a )$, $f(I_p^i )$, $f(I_n^i)\epsilon T$. The loss is minimized by Euclidean distance, which is defined as:

\begin{equation} L_t = \frac{1}{N} \sum_{i=1}^n||f(I_a) - f(I_p^i)||^2_2 - || f(I_a) - f(I_n^i) ||^2_2 > \alpha_t \label{eq6}\end{equation}

\begin{figure*}
\centering
{\includegraphics[scale=0.25]{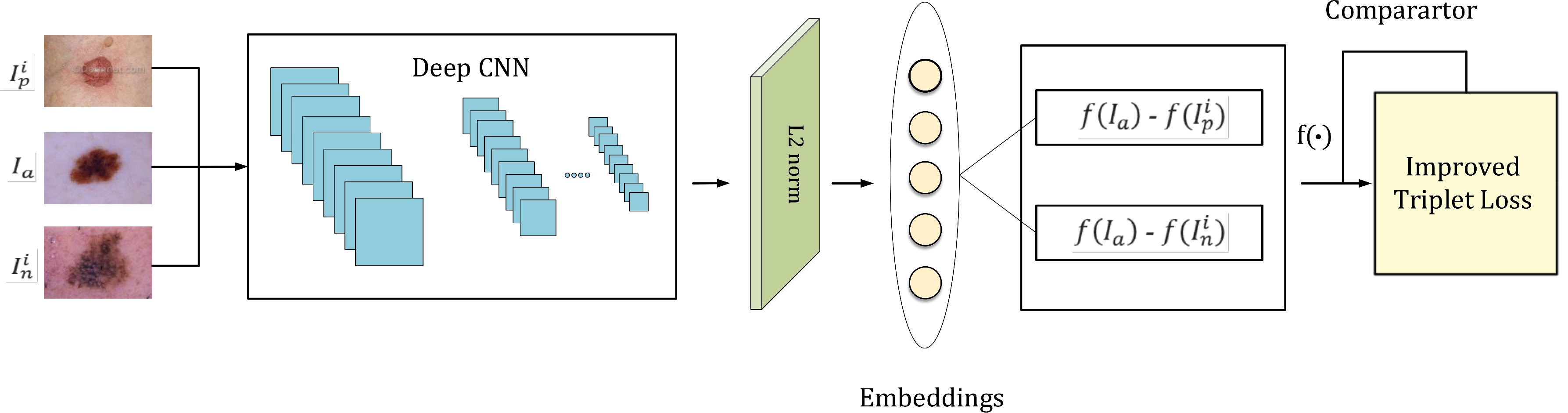}}
\caption{Constrained triplet network structure. The number of batches passes through deep CNN, and skin disease image representation is generated using the L2-normalization layer. In the end, the triplet loss function uses image representation to recognize the pair images (same or different disease).}
\label{Figure4}
\end{figure*}

The TLF needs to create a positive difference between the values of $||f(I_a )-f(I_n^i )||_2^2$, and $||f(I_a )-f(I_p^i )||_2^2$ should be equal to the predefined margin. Although TLF only computes the distance between $f(I_a )$ and $f(I_n^i )$, it does not define how it minimizes the distance between the embedding of the same person; $I_a$ and $I_p^i$. The initial TLF describes the inter-class variation while avoiding the intra-class variation. Consequently, avoiding the intra-class variation may lead to improper distribution of the same class images. The age variation and facial appearance increase the distance between the same class images in a cluster, which leads to inconsistency in recognition consequences. The schematic view of the constrained triplet network is shown in Figure \ref{Figure4}. 

\section{Methodology}
We combine BCNN with a CTN to get the instances or embedding of training images I in d-dimensional Euclidean space $R^d$. We extract the bilinear features $f$ of a specific input image, $I^i$, and then classify the image into $k$ classes using softmax, where $k$ is the number of classes in the dataset. Introduced new feature vector $f_s$ by joining $k$ output values, where $f_s(i)$ is a specific return value of an image classified as a class $L_i$. To minimize the effect of intra-class variations, compute the expectation of the output values of softmax $f_s=\epsilon E(f_s)$ for all images from the same class. The similarity matrix is denoted by $S\epsilon R_(k\times k)$ as Equation \ref{eq7}:

\begin{equation} S_{i,j} = \Xi{\bar{f_s}} \label{eq7}\end{equation}

where $\Xi (\cdot)$ shows the joining of $k\times k$ k-dimensional vectors with $k\times k$ dimensional matrix, $S_(i,j)$ shows the probability of $L_i$ image that is classified as $L_j$.

\begin{figure*}
\centering
{\includegraphics[scale=0.31]{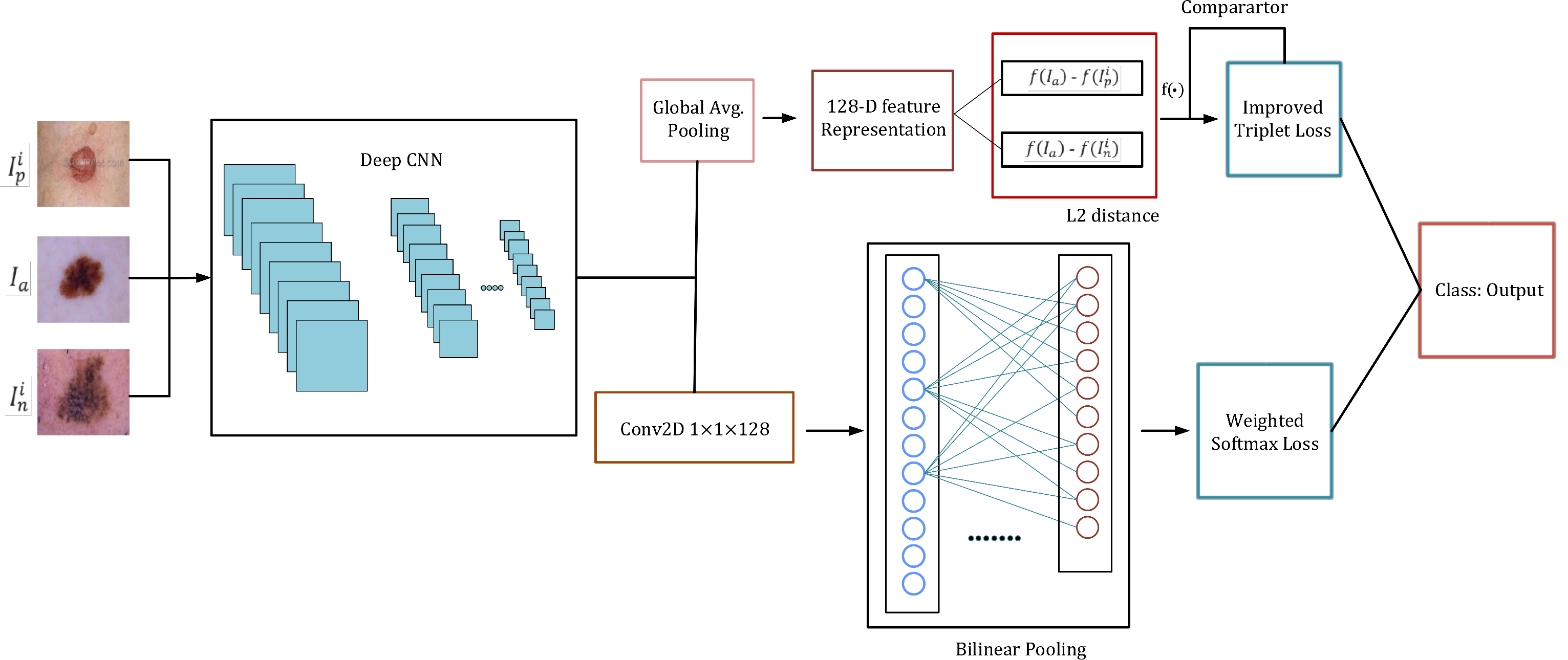}}
\caption{Proposed model architecture.}
\label{Figure5}
\end{figure*}

We proposed similarity matrix-driven based deep metric learning for similarity measurement between very similar class images. A network can correctly increase inter-class similarities and reduce intra-class differences by adaptively sampling the triplets to optimize the especially proposed TLF. However, if we randomly select triplets, most loss function values are 0, affecting convergence during training by backpropagation and trade-off gap between hard triplets and bias in the triplet selection. Therefore, the method requires balancing the trade-off between hard triplets and bias in triplet selection. Thus, we need to compute the maximum and minimum values of input data to reduce the bias in triplet selection. Additionally, suppose we calculate maximum and minimum values for the complete input data. In that case, the poor images can prevail over the hard positive and hard negative skin disease images, which will be the cause of poor training. Therefore, we used an online triplet selection process to compute maximum and minimum values. In the online process, input data is split into small batches (mini-batch)\cite{yu2018correcting}. Consequently, we compute the maximum metric; $||(f(I_a)-f(I_p^t) ||_2^2$ and minimum metric $||(f(I_a)-f(I_n^t)||_2^2$ by using hard-positive $(I_p^t)$, and hard-negative $(I_n^t)$ images from mini-batch. Additionally, similar images are over-sampled to improve the ability of similarity measurement, while other images from different classes are usually sampled to ensure that the method can further distinguish them. The schematic view of the proposed method is shown in Figure \ref{Figure5}.

\subsection{Joint Loss Function}
The joint loss function performs better in feature extraction than the traditional loss function-based methods, and the FC layer weight represents the center of the category. The softmax loss is used to optimize the sample features and close them to the weight vector. Therefore, regularizing the weight vectors is essential to increase the distance between different classes. Let $f(x)$ indicate that feature embedding of image $x$, for a triplet $T=(I_a, I_p^i, I_n^i)\epsilon T_i$, we suppose its distance from any different class image $I_n^i$ is larger than the same class image $I_p^i$ by a pre-defined margin $\mu_1$  as shown in Equation \ref{eq8}:

\begin{equation} ||(f(I_a) - f(I^t_p )||^2_2 + \mu_1 \leq ||(f(I_a) - f(I^t_n )||^2_2 \label{eq8}\end{equation}

where $f(I_a)$, $f(I_p^i)$, $f(I_n^i)\epsilon T$ and $\mu_1 > 0$ is a pre-defined parameter addressing the minimum difference between pair images of the same person and different person, then the triplet loss is defined as:

\begin{equation} L_t = \frac{1}{N} \sum_{i=1}^n [||f(I_a) - f(I_p^i)||^2_2 - || f(I_a) - f(I_n^i) ||^2_2 + \mu_1] \label{eq9}\end{equation}

where $N$ is the total number of input images for selecting triplets. According to the discriminative feature learning algorithms using CNN, the softmax loss is commonly used to tackle the traditional classification constraints and correctly classify or differentiate the different class images. Although intra-class differences are not fixed, these differences are crucial to locating visually and semantically similar instances. The TLF is added with a new constraint to reduce the distance between positive pairs; $I_a$, $I_p^i$ within the same class by less than $\mu_2$ as shown in Equation \ref{eq10}:

\begin{equation} ||(f(I_a) - f(I^t_p )||^2_2 \leq \mu_2 \label{eq10}\end{equation}

Hence, the improved triplet-loss function is expressed as Equation \ref{eq11}:

\begin{equation} L_{Triplet} = b[|| f(I_a) - f(I_p^i) ||^2_2 - \mu_2 ]  +  \frac{1}{N} \sum_{i=1}^n [||f(I_a) \\ - f(I_p^i)||^2_2 -  || f(I_a) - f(I_n^i) ||^2_2 + \mu_1] \label{eq11}\end{equation}

where $\mu_2 > 0$ is a pre-defined margin addressing the maximum difference between a pair of images of the same and a different class. The convergence speed of improved TLF can slow with increasing the size of the input as compared to traditional TLF, and easier to overfit as well. Thus, the triplet and softmax loss must be optimized jointly to improve convergence speed. The classic softmax loss function equally treats all different class images; hence, to further improve the ability to differentiate different class images, the softmax loss function will improve to make the model stricter for the wrong classification. Equation \ref{eq12} shows the wrong-classification probability $P_i$ based on the similarity matrix $S$:

\begin{equation} P_i = \sum_{j=1,i \neq j}^k S_{i,j} \label{eq12}\end{equation}

Therefore, the weighted softmax loss is calculated by Equation \ref{eq13}:

\begin{equation} L_{softmax} = \frac{1}{N} \sum_{i=1}^N -P_{i} * log({f_s}(i)) \label{eq13}\end{equation}

where $\Xi f_s(i)$  is the output of the image and is classified as a class labelled $L_i$. The triplet-loss layer computes the similarity loss using feature embedding of triplets, and the softmax loss layer uses the same feature embedding of a triplet to compute classification error. Subsequently, we compute the joint loss of these two loss functions by Equation \ref{eq14}:

\begin{equation} L_{joint} = \alpha_t L_{softmax} + (1 - \alpha_t) L_{Triplet} \label{eq14}\end{equation}

where $\alpha _t$ is the predefined margin that controls the trade-off between these losses, the Joint loss function optimization algorithm is presented in Algorithm $1$.

\section{Dataset and Evaluation Metrics}
\subsection{Dataset}
We trained and evaluated the proposed method on the benchmark ISIC2019 dataset\cite{kassem2020skin}. The publically available ISIC2019 contains images from HAM10000\cite{tschandl2018ham10000} and BCN-20000. The HAM10000 dataset contains images of size 600×450 that were the same as ISIC 2018, centred, and then cropped the lesion. BCN-20000 dataset contains images of size 1024×1024 that include an additional unknown class in the test set, which was not presented in the training dataset. The dataset has eight classes, namely Melanoma (MEL), Melanocytic Nevus (NV), Basal Cell Carcinoma (BCC), Actinic Keratosis (AKIEC), Benign Keratosis (BKL), Dermatofibroma (DF), Vascular Lesion (VASC), and Squamous Cell Carcinoma (SCC). The dataset consists of a total of 25331 images with different numbers of all classes. One of the most challenging factors for classifying the dataset's images is an imbalanced number of images in each class. Thus, we preprocessed these samples using several arguments, i.e., rescale, rotation with a range of 0.3, horizontal flip, and zoom with a range of 0.3. Afterwards, the dataset was augmented to 25,600 samples with the same number of images in every class. We used 80\% of the images for training, 20\% for testing, and 10\% of training data are used for validation.
\subsection{Evaluation Metrics and Algorithm}
For model performance evaluation, we used the following metrics:

\begin{equation} Accuracy = \frac{TP + TN}{TP+ FP + TN + FN} \label{eq15}\end{equation}

Sensitivity (Recall) shows the positively predicted rate with respect to wrongly identified cases as negative ailments:

\begin{equation} Sensitivity (Recall) = \frac{TP}{TP + FN} \label{eq16}\end{equation}

Specificity shows the true negative rate, i.e., the true negatively identified rate with respect to the wrongly identified case as negative ailments.

\begin{equation} Specificity = \frac{TN}{TN + FP} \label{eq17}\end{equation}

where TP is correctly identified as a positive case, TN is correctly identified as a negative case, FP is wrongly identified as a positive case, and FN is wrongly identified as a negative case.

We also used AUC (Area Under Curve) to graphically demonstrate the classification skill of the proposed method, which is plotted TPR (true positive rate) against FPR (false positive rate). TPR and FPR are computed using the following equations:

\begin{equation} TPR = \frac{TP}{(TP + FN)} \label{eq18}\end{equation}
\begin{equation} FPR = \frac{FP}{(FP + TN)} \label{eq19}\end{equation}

The higher value of AUC implies better results (See Figure \ref{Figure6}).

\begin{algorithm}[hbt!]
\caption{Joint-Loss Function Optimization Algorithm}
\label{alg:joint_loss_optimization}
\begin{algorithmic}[1]
\STATE \textbf{Input:} Training set $I^i$, initialized parameter $W \in \mathbb{R}^{d \times n}$ in the convolution layers, learning rate $\eta$, hyperparameters $\alpha_t$, $\beta$
\STATE \textbf{Output:} Updated parameter $W$
\STATE $t \gets 0$
\WHILE{not converged}
    \STATE $t \gets t + 1$
    \STATE Compute the first loss $L_t = E^2(w^T w)$
    \STATE Compute the second loss $L_c = E((w^T w)^2)$
    \STATE Compute the joint loss $L_{\text{joint}} = L_* + \alpha_t L_c + \beta L_t$
    \STATE Calculate the backpropagation process $\frac{{\partial L_{\text{joint}}}}{\partial w^{t}_i}$ for each parameter by $\frac{{\partial L_{\text{joint}}}}{\partial w^{t}_i} = \frac{{\partial L_*}}{\partial w^{t}_i} + \frac{{\partial L_c}}{\partial w^{t}_i} + \frac{{\partial L_t}}{\partial w^{t}_i}$
    \STATE Update the parameter $w_i^{t + 1} = w_i^t - \eta \frac{{\partial L_{\text{joint}}}}{\partial w^{t}_i}$
\ENDWHILE
\RETURN Updated parameter $W$
\end{algorithmic}
\end{algorithm}

\section{Experiments and Results}

\subsection{Model Architecture}
The model architecture consists of CTN with BCNN. The input images of sizes (150, 150, 3) were used. After feature extraction by Xception, we retrieve a feature map of size (14, 14, 1024). We already mentioned above that all branches of Xception are structured on their characteristics. The global average pooling layer is added to enormously increase the localization ability of the deep CNN, even with training on image-level labels\cite{zhou2016learning}. After global-average pooling, we retrieve a 128-D vector and then generate distance representation using feature embeddings; although the distance will be either negative or positive, the selected pair of sample images are predicted as different identities. The other part of the Xception model reduces the filter dimension of size (1, 1, 128), and the output directly connects with the bilinear vector. To improve the performance of the method, bilinear pooling flattens the bilinear feature, then the signed square-root function is applied to the flattened feature, and finally, applies L2-normalization on the result.

\subsection{Input and Network Settings}
We pre-process the input data with some arguments such as zoom, flip horizontal, zoom of range 0.3, rotation with range 0.3, set data format as channel last, and generate a floating array of input images. After pre-processing, we normalized the input floating values $(-1, +1)$. We divide our dataset into seven parts to compute $L_2$ distance; the first five parts are used in training, and the last two are used for testing and validation. In order to improve the efficiency of the model and achieve a good result in a few iterations, we used the parameters of the pre-trained model on ImageNet\cite{krizhevsky2012imagenet} by initializing the convolutional and softmax layers for feature extraction. In our experiments, we first use the Adam optimizer to train the branch of the bilinear model, and then we use SGD with a learning rate of 0.0001 to train two branches together. The network achieved an effective result by following this approach. Additionally, we used Dropout regularization of values $0.25-0.5$ to increase the generalization ability and minimize the over-fitting problem. The first half of end-to-end training starts with the convolutional and pooling layers. Thus, in the latter half of the model, we used gradient values for training. Assume that the feature extractor; $f_A$  and $f_B$  get $A$ and $AT$ at each location $l$. The bilinear pooled feature is $A\times  AT$. Eventually, we used the chain rule to obtain the gradient of the loss function at the output position of the network to complete the training. The branch of BCNN predicts the input image, while the CTN recognizes the pair images (same or different).

\subsection{Training Loss and Accuracy}
\begin{figure*}
\centering
\includegraphics[scale=0.60]{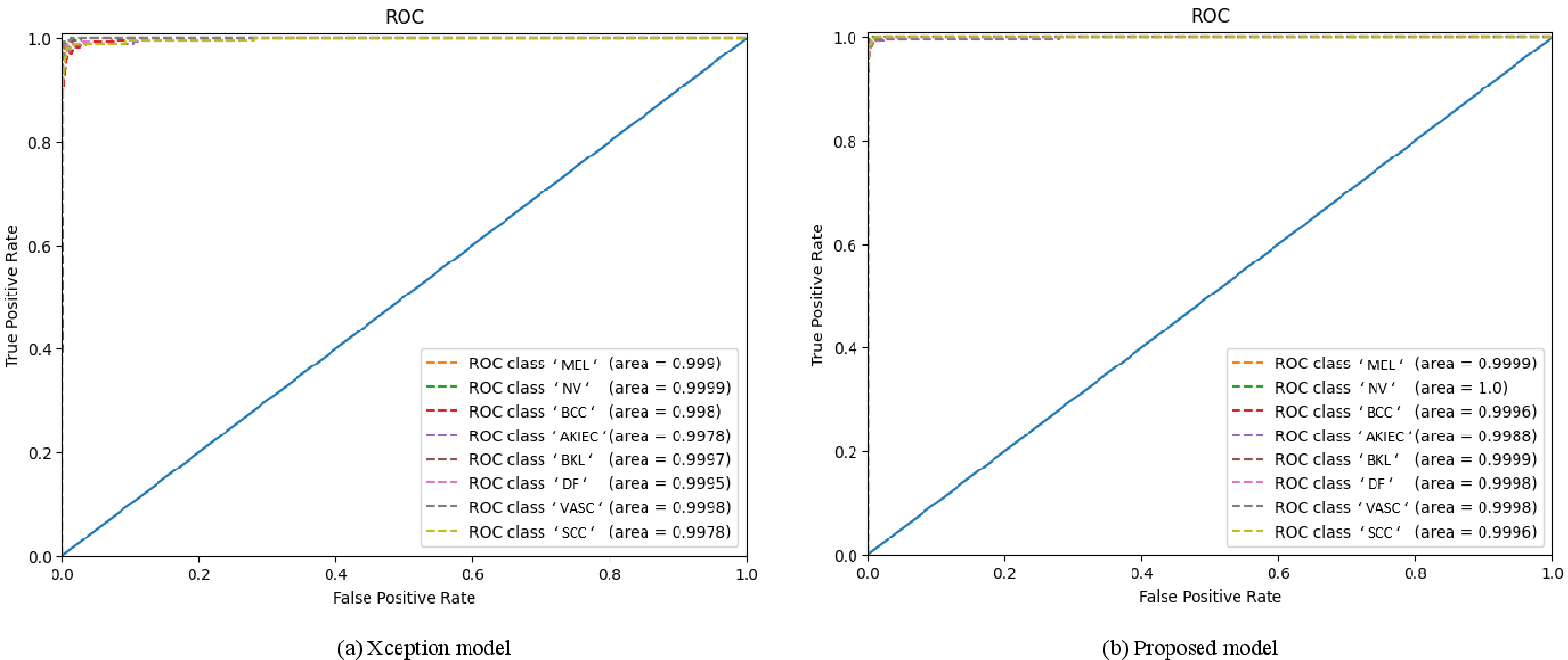}
\caption{Figure shows the Area Under Curve (AUC) values for all categories of skin disease individually by (a) the Xception model and (b) the proposed model.}
\label{Figure6}
\end{figure*}

\begin{figure*}
\centering
\includegraphics[scale=0.85]{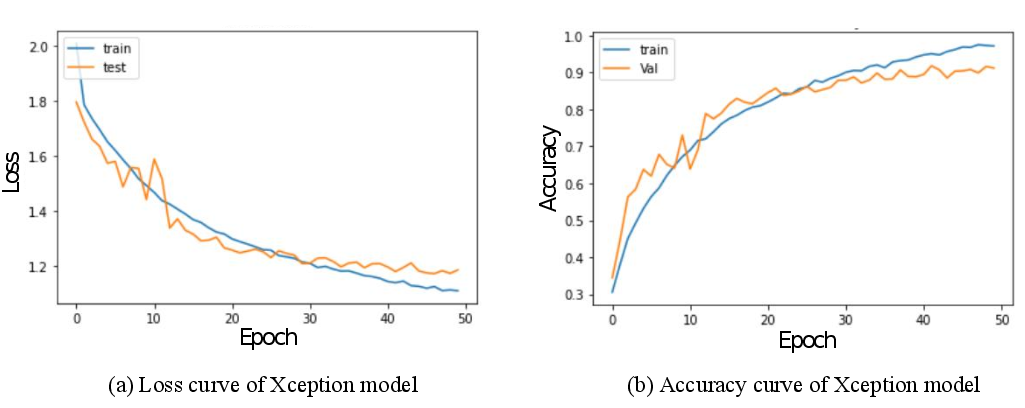}
\caption{Trend of loss and accuracy with the number of epochs for training and validation data using Xception model.}
\label{Figure7}
\end{figure*}

\begin{figure*}
\centering
\includegraphics[scale=0.85]{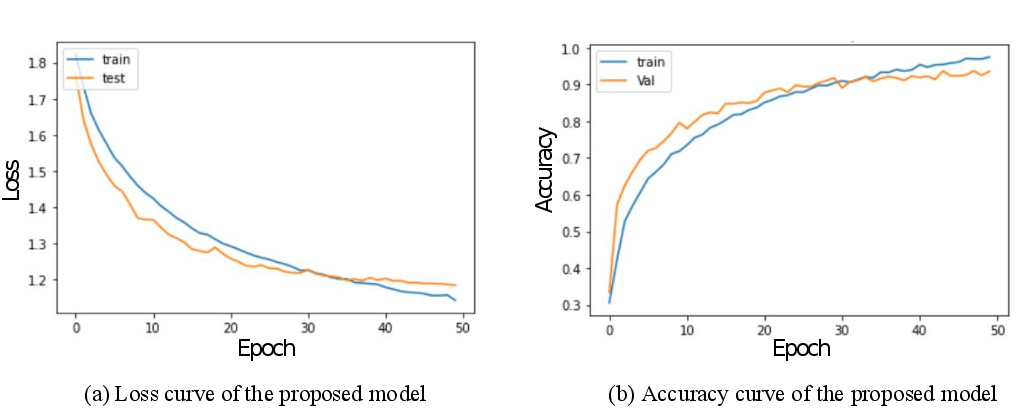}
\caption{ Trend of loss and accuracy with the number of epochs for training and validation data of the proposed model.}
\label{Figure8}
\end{figure*}

\begin{figure*}
\centering
\includegraphics[scale=0.60]{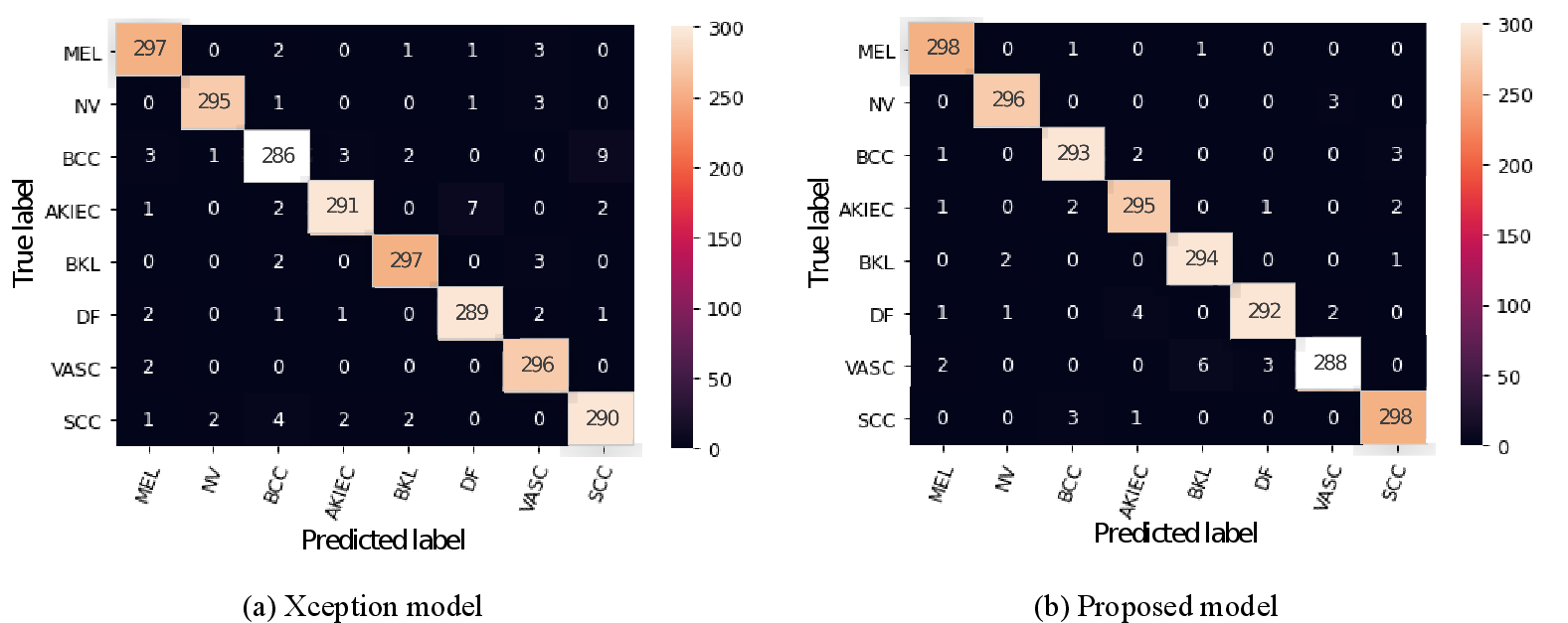}
\caption{Confusion matrix of (a) Xception model and (b) proposed methods shows the prediction values for all categories of skin disease images. Each column of the matrix indicates the instances in a predicted class, and each row indicates the instances in an actual class.}
\label{Figure9}
\end{figure*}

To evaluate the performance, we used the evaluation matrix which is described in section $5$. Furthermore, we used AUC to graphically demonstrate the classification skill of the proposed method as shown in Figure \ref{Figure6}, which is plotted by using TPR against FPR. We have plotted the loss, and accuracy (i.e., training and validation) against the number of epochs for Xception and proposed methods, which are shown in Figure \ref{Figure7}, and \ref{Figure8}, respectively. Moreover, the confusion matrices of the proposed networks are shown in Figure \ref{Figure9}. The row values of confusion metrics denote the corresponding true label, and column indices denote the corresponding predicted labels. The value that appears in each cell of the confusion matrices shows the prediction labels. Here, we can see that the diagonal cell of confusion metrics acquires a high level of prediction, which indicates a low error rate (high probability of accurate prediction) for each category of skin diseases. To test the efficiency of the proposed method, we computed some evaluation factors, namely, sensitivity (recall) and specificity, using confusion metrics, which are demonstrated in Table \ref{table1}. Additionally, we computed AUC (true accept rate against false accept rate) values for all categories of skin diseases.

\begin{table*}[!ht]
\caption{Classification performance of the proposed method. A total of 300 skin disease images from each class are used for validation.}
\begin{center}
\setlength{\tabcolsep}{2pt}
\begin{tabular}{|c|ccc|ccc|c|}
\hline
\rule{0pt}{1.0\normalbaselineskip}
Categories & \multicolumn{3}{|c|}{Xception + B-CNN} & \multicolumn{3}{|c|}{Proposed Method} & Samples no. \\[5pt]
\hline
{} & \rule{0pt}{1.0\normalbaselineskip} Accuracy(\%) & Sensitivity(\%) & Specificity(\%) & Accuracy(\%) & Sensitivity(\%) & Specificity(\%) &  \\[5pt]
\hline
\newcommand\Tstrut{\rule{0pt}{2.6ex}}
MEL & - & 92.65 & 100 & - & 94.55 & 100 & 300 \\[5pt]
NV & - & 90.71 & 95.87 & - & 92.72 & 99.62 & 300 \\[5pt]
BCC & - & 92.34 & 93.61 & - & 93.66 & 98.96 & 300 \\[5pt]
AKIEC & - & 92.10 & 99.29 & - & 94.92 & 100 & 300 \\[5pt]
BKL & - & 91.53 & 97.23 & - & 92.74 & 100 & 300 \\[5pt]
DF & - & 91.48 & 99.62 & - & 93.89 & 99.43 & 300 \\[5pt]
VASC & - & 94.21 & 100 & - & 94.57 & 99.45 & 300 \\[5pt]
SCC & - & 93.56 & 99.67 & - & 97.23 & 100 & 300 \\[5pt]
\hline
\rule{0pt}{1.0\normalbaselineskip}
Average & 91.58 & 92.32 & 98.16 & 93.72 & 94.28 & 99.68 & 300 \\[5pt]
\hline
\end{tabular}
\end{center}
\label{table1}
\end{table*}

\subsection{Effect of Alpha}

\begin{figure*}
\centering
{\includegraphics[scale=0.35]{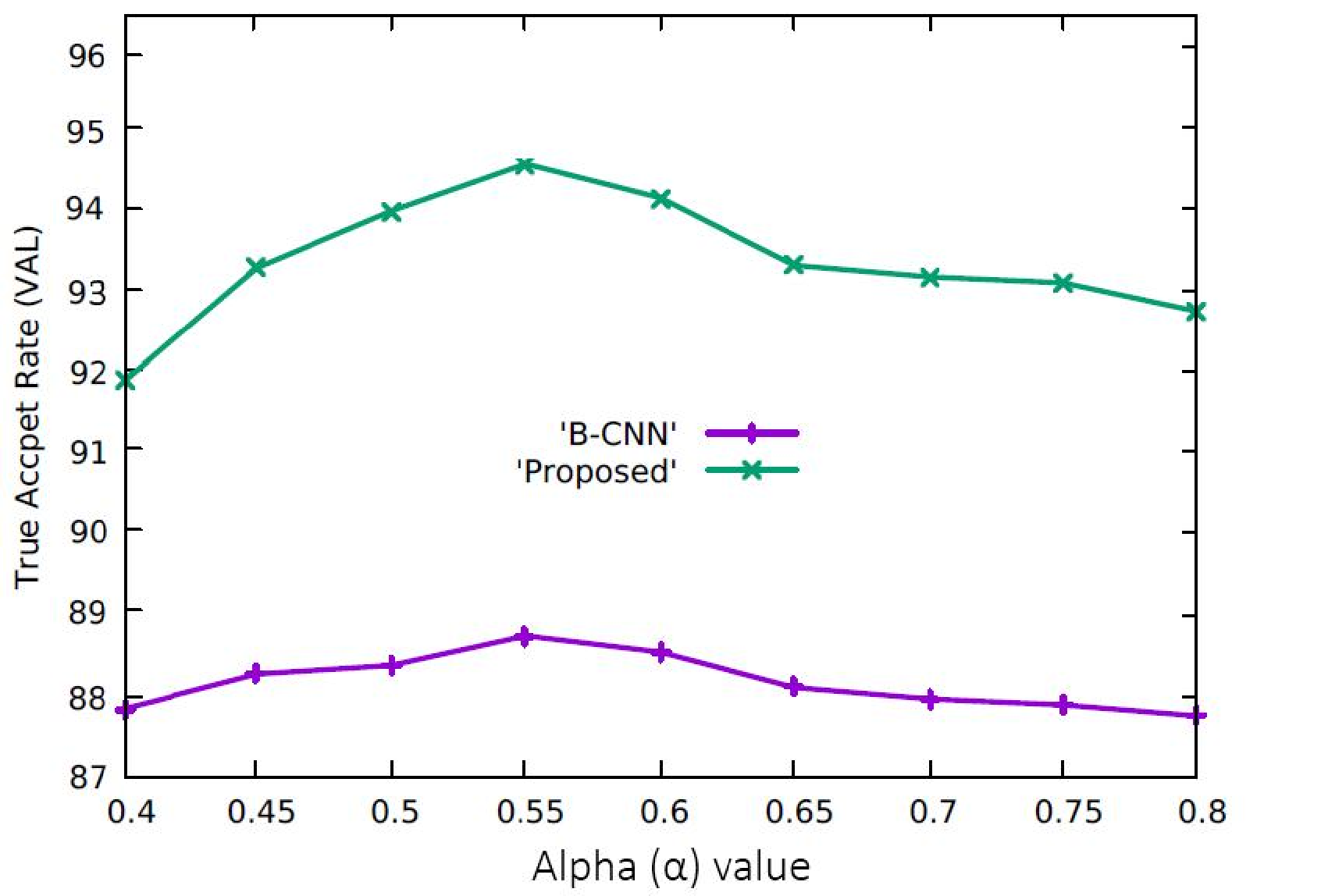}}
\caption{Model accuracy at different Alpha $(\alpha)$ values.}
\label{Figure10}
\end{figure*}

Four parameters are used in our framework such as $\mu_1$, $\mu_2$, $\alpha_t$, $\beta$, where $\mu_1 > 0$ is a pre-defined parameter addressing the minimum difference between pair images of the same person and a different person, and $\mu_2$ denotes the new constraint added by the triplet loss function to reduce the distance between positive pairs; $I_a$, $I_p^i$ within the same class by less than $\mu_1$, $\alpha_t$ is a pre-defined margin that is imposed between images from different categories, and $\beta$ is the learning rate. In all our experiments, parameters $\mu_1$, $\mu_2$, and $\beta$ were set as a specific value. 
Consequently, we only changed the parameter $\alpha_t$ in training to balance the losses and measure the impact on results with the convergence rate. Figure \ref{Figure10} shows that the accuracy improved with the threshold and started to drop when the threshold was taken to $0.55$. So, we could assume that we achieved the best result on a threshold value of $0.55$.

\subsection{Pair Testing}

Online testing is enabled to monitor every epoch. An evaluation is performed to make the current model good at the end of every epoch. The validation process is straightforward but significant and evaluated on the pair image selected from the validation data. At the validation, the current model feds to each pair of images. The loss layer is removed because back-propagation is not required to adjust the model weights. Therefore, only forward propagation needs to generate feature embeddings at this stage. Meanwhile, both the representations of feature embeddings are generated to distinguish both the images using Euclidean distance, which is defined as:

\begin{equation} dist_{(x, y)} = \sqrt{(x_{1} - y_{1})^{2} + (x_{2} - y_{2})^{2} + ... + (x_{n} - y_{n})^{2}} \label{eq20}\end{equation}

where $(x_i,y_i)$ is the $i^th$ pair of images of the validation data, and n is the total number of pairs in the validation dataset. The number of pair images for validation is randomly selected for evaluation. However, we have selected 60\% of image pairs of the same class and 40\% from different classes to perform a fair evaluation. We used 10-fold of 600 images in our experiment to obtain effective results. Moreover, we reported the average accuracy on splits with cross-validation that followed standard validation protocols. Each fold has the same number of input samples. In each iteration, the nine folds are used for training, and the last fold is used to calculate accuracy. Calculating the distance between a pair of images requires a threshold value to separate them. For this procedure, the best threshold value is obtained and then used on the training folds, and the same threshold value is used on the test folds.

\subsection{Comparative Study}
\begin{figure*}
\centering
{\includegraphics[scale=1.0]{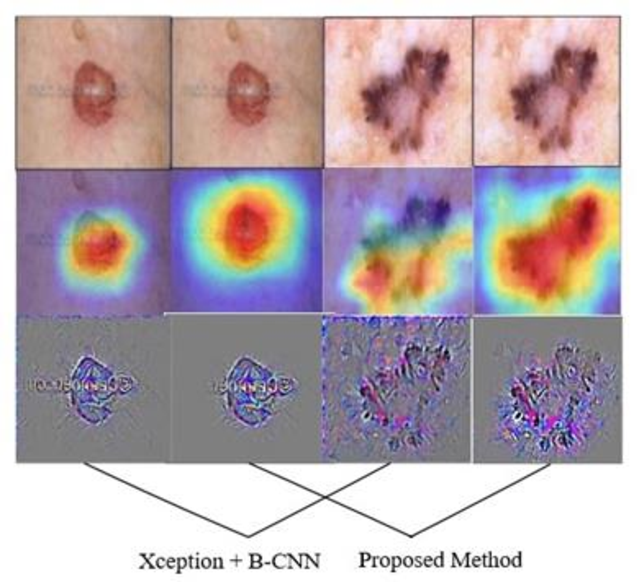}}
\caption{Comparative gradient-weighted class activation mapping (Grad-CAM) and Guided Grad-CAM visualizations of Xception+bilinear mode with the proposed method. The second row is the Grad-CAM results highlighting the important regions in red; the third row is the high-resolution class-discriminative visualizations of Guided Grad-CAM.}
\label{Figure11}
\end{figure*}

CNN-based methods use Gradient-weighted class activation mapping (Grad-CAM) for visualizing the essential parts of the input from a predictive viewpoint to enhance the transparency\cite{selvaraju2017grad} as well as Saliency feature maps are used to show the impact of each pixel in the image on the results\cite{simonyan2013deep}. We visualize Grad-CAM explanations and saliency feature maps of dense layers by integrating the CTN to understand our architecture better, and representative results are shown in Figure \ref{Figure11}, which shows the improvement of classification results. All the image components, such as marked degree, color, texture, etc., validate the method by effectively extracting the features to distinguish among the different diseases. Several DL-based state-of-the-art classification methods have been regularly renovated accuracy in recent years.
In order to examine the concern of the proposed method, we compared the results (i.e., accuracy, specificity, and sensitivity) of the proposed method with some existing methods on the ISIC2019 benchmark dataset are shown in Table \ref{table2}. The accuracy and sensitivity of the proposed method exceed all comparable methods\cite{nikolaou2014emerging, esteva2017dermatologist, amelard2014high, trigueros2018enhancing, kassem2020skin, capdehourat2011toward, zaqout2019diagnosis, abbas2013pattern, blum2004digital, chakraborty2017image, brinker2018skin, khan2018implementation, yang2018classification}, even though the sensitivity of Amelard et al.\cite{amelard2014high} and Esteva et al.\cite{esteva2017dermatologist} are better than the proposed method, but its specificity is very low. These results indicate that discriminative feature learning is effective for many basic CNN architectures, and our method learns more effectively than state-of-the-art methods.

\begin{table*}[!ht]
\center
\setlength{\tabcolsep}{5pt}
\caption{Comparison of the proposed method with some traditional methods. The value in bold indicates the best result.}
\begin{tabular}{|c|ccc|}
\hline
\rule{0pt}{1.0\normalbaselineskip}
Method & Accuracy ( \%) & Sensitivity (\%) & Specificity (\%) \\[5pt]
\hline
\rule{0pt}{1.0\normalbaselineskip}
MobileNetV2-LSTM  \cite{nikolaou2014emerging} & 85.34 & 88.24 & 92.00 \\[5pt]
Esteva et al.  \cite{esteva2017dermatologist} & 72.1 & 96 &  \\[5pt]
Amelard et al.  \cite{amelard2014high} & 81.17 & \bf 96.64 & 65.06 \\[5pt]
Daniel et al.  \cite{trigueros2018enhancing} &  & 78.43 & 97.87 \\[5pt]
Kassem et al.  \cite{kassem2020skin} & 81.0 & 74.0 & 84.0 \\[5pt]
German et al.  \cite{capdehourat2011toward} &  & 90 & 77 \\[5pt]
Zaqout et al.  \cite{zaqout2019diagnosis} & 90.0 & 85.0 & 92.22 \\[5pt]
Qaisar et al.  \cite{abbas2013pattern} &  & 89.28 & 93.75 \\[5pt]
MobileNetV2  \cite{blum2004digital} & 84.00 & 86.41 & 90.00 \\[5pt]
Chakraborty et al.  \cite{chakraborty2017image} & 90.56 & 88.26 & 93.64 \\[5pt]
Brinker et al.  \cite{brinker2018skin} & 76.90 & 89.40 & 64.40 \\[5pt]
MKhan et al.  \cite{khan2018implementation} & 88.20 & 88.50 & 91.0 \\[5pt]
Yang et al.  \cite{yang2018classification} & 83.0 & 60.70 & 88.40 \\[5pt]
Proposed method & \bf 93.72 & 93.86 & \bf 99.10 \\[5pt]
\hline
\end{tabular}
\label{table2}
\end{table*}

\section{Conclusion}
In this paper, we proposed a method based on BCNN with a CTN for skin disease classification. Our proposed method trains in an end-to-end way to effectively increase the distance between inter-class disease image features, which maintains the intra-class image representations closer to improve the classification accuracy. We employed our method on datasets consisting of skin disease images in six categories and compared the results with some traditional methods that confirm our method achieves better classification accuracy than conventional methods. Our proposed method achieved 93.72\% classification accuracy. Our future work will mainly focus on two issues. Firstly, we will be integrating losses in different branches of a model, and secondly, we will consider using the attention technique with the fine-grained process.

\bibliographystyle{unsrt}  
\bibliography{arxiv}

\end{document}